\documentclass[sigconf]{acmart}
\usepackage[noend]{algpseudocode}
\usepackage{algorithm,algorithmicx}

\copyrightyear{2022}
\acmYear{2022}
\setcopyright{acmcopyright}
\acmConference[SBST'22]{The 15th Search-Based Software Testing Workshop}{May 9, 2022}{Pittsburgh, PA, USA}
\acmBooktitle{The 15th Search-Based Software Testing Workshop (SBST'22), May 9, 2022, Pittsburgh, PA, USA}
\acmPrice{15.00}
\acmDOI{10.1145/3526072.3527535}
\acmISBN{978-1-4503-9318-8/22/05}

\title{WOGAN at the SBST 2022 CPS Tool Competition}

\author{Jarkko Peltomäki}
\affiliation{%
  \institution{Åbo Akademi University}
  \city{Turku}
  \country{Finland}}
\email{jarkko.peltomaki@abo.fi}
\author{Frankie Spencer}
\affiliation{%
  \institution{Åbo Akademi University}
  \city{Turku}
  \country{Finland}}
\email{frankie.spencer@abo.fi}
\author{Ivan Porres}
\affiliation{%
  \institution{Åbo Akademi University}
  \city{Turku}
  \country{Finland}}
\email{ivan.porres@abo.fi}

\begin{abstract}
  WOGAN is an online test generation algorithm based on Wasserstein generative adversarial networks. In this note, we
  present how WOGAN works and summarize its performance in the SBST 2022 CPS tool competition concerning the AI of a
  self-driving car.
\end{abstract}

\begin{document}

\maketitle

\section{SBST 2022 CPS Tool Competition}
The SBST 2022 CPS tool competition is concerned with finding road scenarios that cause the AI of the BeamNG.tech
driving simulator to drive a car out of its designated lane. The organizers of the competition performed two
experiments in which the competition entries were compared according to three metrics: efficiency, effectiveness, and
failure-inducing test diversity. The BeamNG.AI was used in the first experiment while the DAVE-2 AI was used in the
second. Complete details on the simulator, road scenarios, the AIs, and experiment results are found in the
competition report \cite{SBST-toolcomp22}.

\section{About WOGAN}
Here we provide a brief explanation how WOGAN works. A more complete description is found in \cite{sbst_long}.

We consider an input road to the BeamNG.tech simulator as a \emph{test}, and we call the output of the test its
\emph{fitness}. We chose the fitness to be the maximum percentage of the body of the car that is out of the boundaries
of its lane during the simulation (in short $\mathsf{BOLP}$). In the experiments of the competition report
\cite{SBST-toolcomp22}, tests were considered failed if $\mathsf{BOLP}$ was over $0.85$ (BeamNG.AI) or $0.10$ (DAVE-2).

The central idea is to use a generative machine learning model to generate tests with high fitness. We chose to train a
Wasserstein generative adversarial network (WGAN) which is capable of producing diverse samples from its target
distribution \cite{wgan}. Since we lack training data in advance, we need to train the WGAN online. Our approach is to
first do a random search to obtain an initial training data for the WGAN and then augment this training data by
executing tests generated by the WGAN which are estimated to have high fitness.

The initial random search produces a training data $(t_i, f_i)$, $i = 1, \ldots, N$, of tests $t_i$ and fitnesses
$f_i$. The WGAN, say $G$, is trained with tests of high fitness (what ``high'' means is explained in \cite{sbst_long}).
In order to find a new candidate test, the WGAN $G$ is sampled for new tests. In order to choose the best candidate
test, we employ an analyzer $A$ which is a neural network trained to learn the map from tests to fitnesses (the
collected training data enables us to do so). Thus $A$ acts as a surrogate model for the simulator, and by using it we
avoid costly executions on the simulator. We select the candidate test $t$ with the highest estimated fitness and
execute it on the simulator to find its true fitness $f$. Finally we add $(t, f)$ to the training data, retrain the
models, and repeat the preceding procedure until the test budget is exhausted. The tests of the final training data can
be considered as a test suite for the simulator.

Intuitively our WOGAN algorithm should be able to find tests with high fitness. As more training data is available, the
generator $G$ should be more capable of generating high-fitness tests and the analyzer $A$ should be more accurate in
assessing test fitness. The experiments described in \cite{sbst_long} and the experiments of the competition report
\cite{SBST-toolcomp22} experimentally validate this intuition.

%We remark that online training of generative models is a widely open problem. We take the experimental results as an
%indication that the ideas presented above could be a viable approach to solve this important problem.

Using WOGAN requires choosing hyperparameters. For the tool competition, we used the same hyperparameters as described
in \cite[Sec.~3.1]{sbst_long}. Most importantly, we used $20 \%$ of the execution budget for random search, and we
always generate roads defined by exactly $6$ points (this is an arbitrary decision). For more on the input
representation and normalization, see \cite[Sec.~2.1]{sbst_long}.

\section{Results and Their Interpretation}
Here we summarize the results of the experiments of \cite{SBST-toolcomp22}. The metrics considered here are efficiency,
effectiveness, and failure-inducing test diversity; see \cite{SBST-toolcomp22} for their more elaborate definitions.\\

\noindent
\textbf{Efficiency.}
As written in \cite{SBST-toolcomp22}, test generation efficiency is measured as the time spent to generate tests. In
the experiments, two time budgets are specified: generation budget ($1 \, \textup{h}$) and simulation budget
($2 \, \textup{h}$). The latter is the time used for running the simulator while the generation budget accounts for the
remaining time used.

\begin{figure}
  \includegraphics[width=\linewidth]{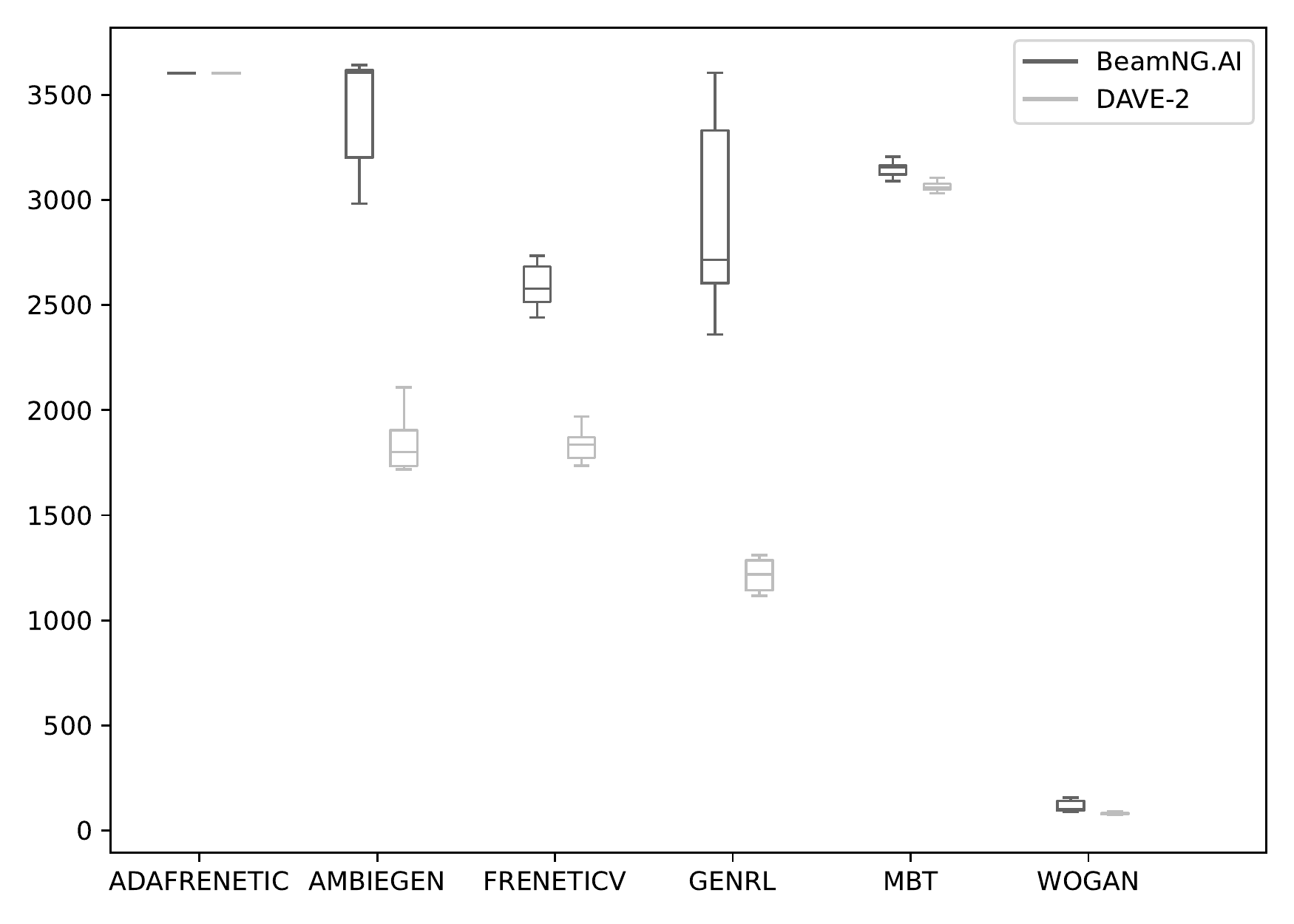}
  \caption{\textmd{Efficiency (in seconds)}}\label{fig:eff}
  \vspace{-1.4em}
\end{figure}

WOGAN is the most efficient tool in the above sense: it uses least time to propose a new test \cite{SBST-toolcomp22}.
In fact, it uses one magnitude less time than the other tools; see \autoref{fig:eff}. We believe that this is mainly
due to the fact that WOGAN is not a traditional SBSE algorithm. The actual search for a new test occurs when the
weights of the neural networks are adjusted. After this, candidate tests are sampled and analyzed which amounts to few
forward passes of neural networks. These operations run efficiently on modern hardware especially since we have very
little training data and small networks.

Since WOGAN does not use most of the generation budget, it would make sense to use more time for generation. More
complicated models or ensembles of models could be trained to improve the metrics. In addition, more domain knowledge
could be utilized.\\

\noindent
\textbf{Effectiveness.}
Not all input roads are valid, so a good tool needs to learn to avoid producing invalid roads. Effectiveness is 
defined as the ratio of valid tests over all generated tests \cite{SBST-toolcomp22}.

WOGAN generates many invalid tests: roughly $50 \%$ of the tests could not be executed \cite{SBST-toolcomp22}. WOGAN
has no built-in notion of a valid test, so the neural networks used need to learn this by trial and error. There is no
smooth function measuring how close a road is being valid, so learning the notion validity accurately with a small
training data can be challenging.

The situation could perhaps be improved by introducing an additional validator classifier model, but we did not
experiment with this. We do not view this notion of effectiveness particularly important as an efficient validator is
provided in the simulation suite meaning WOGAN could validate a candidate road internally before proposing it as a test
being simulated. This way it could achieve a perfect effectiveness with $0$ invalid tests. This was also noted in the
competition note for Frenetic 2021 \cite{frenetic}.\\

\begin{figure}
  \includegraphics[width=\linewidth]{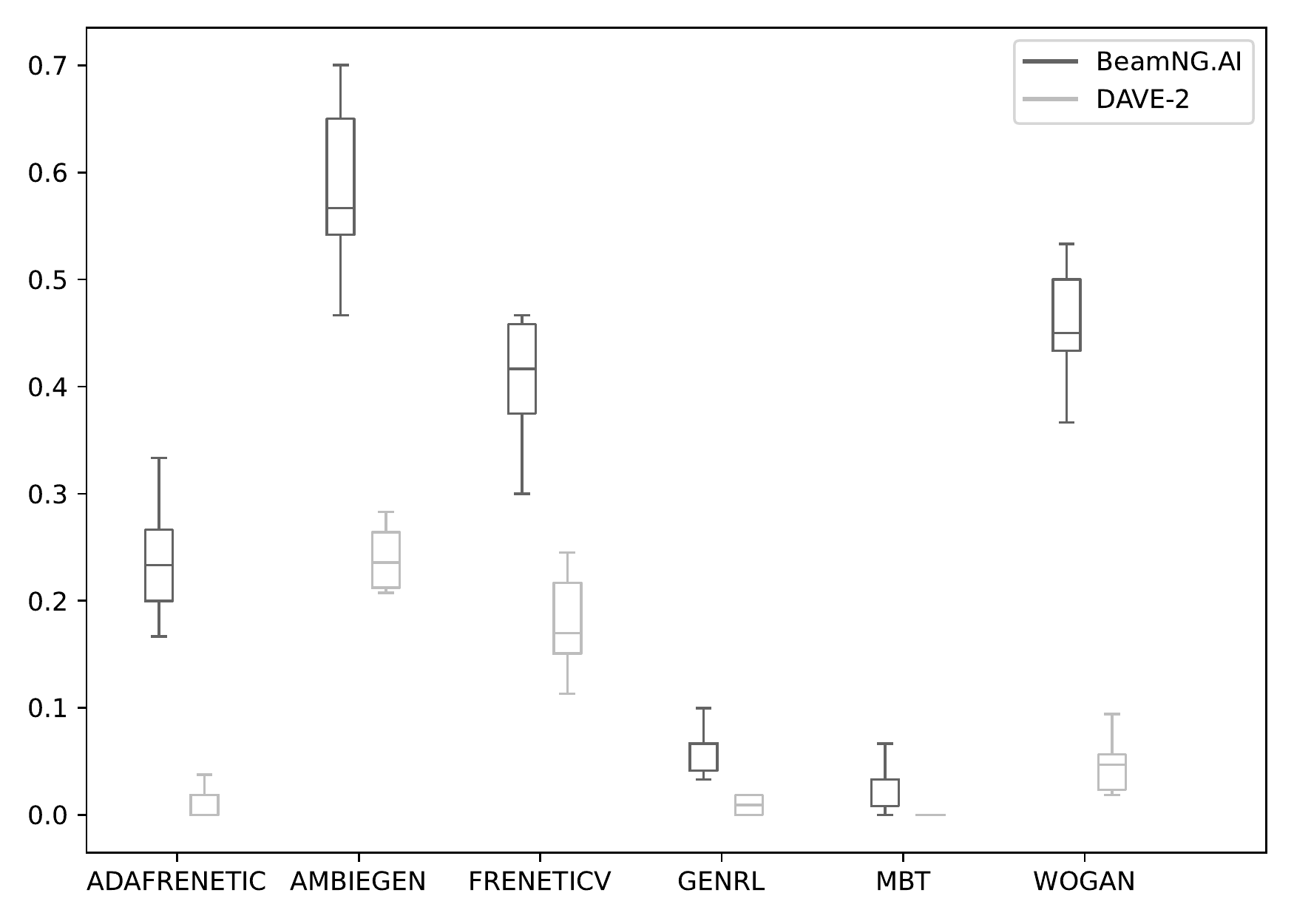}
  \caption{\textmd{Relative map coverage}}\label{fig:diversity}
  \vspace{-1.4em}
\end{figure}

\noindent
\textbf{Failure-Inducing Test Diversity.}
In the BeamnNG.AI experiment WOGAN generated a large number of failed tests: on average $330.3$ (SD $55.8$) over $10$
repetitions whereas the next best tool, AMBIEGEN, generated on average $90.4$ (SD $12.0$) failed tests. The tests
generated by WOGAN were however not as diverse as those generated by AMBIEGEN (see \autoref{fig:diversity}), so WOGAN
was ranked second in the failure-inducing test diversity metric \cite{SBST-toolcomp22}.

In the DAVE-2 experiment, WOGAN did not fare that well. It was ranked third by failure-inducing test diversity, but its
performance was considerably lower than that of AMBIEGEN and FRENETICV; see \autoref{fig:diversity}. WOGAN was able
to generate on average $3.1$ (SD $1.3$) failed tests over $10$ repetitions. AMBIEGEN and FRENETICV respectively achieved the
means $15.3$ (SD $6.3$) and $11.1$ (SD $4.5$).

We believe the reason for WOGAN's worse performance in the DAVE-2 experiment is that the $\mathsf{BOLP}$ output of the
simulator is not sensitive enough. We have observed that a randomly chosen road often has a very low $\mathsf{BOLP}$
value in the DAVE-2 setting, and the neural networks could have trouble learning with so homogeneous data. We believe
we should have considered the distance of the center of the car to the edges of the lane. This distance yields
information even when $\mathsf{BOLP}$ is $0$. One additional factor is hyperparameter tuning: we mainly used the
BeamNG.AI for development. Perhaps our hyperparameters for BeamNG.AI are subpar for DAVE-2.

We are satisfied with the diversity of roads generated by WOGAN. We relied solely on the ability of WGAN's to produce
varied samples, and we did not utilize any domain-specific knowledge. Moreover, we worked under the assumption that the
road diversity metric was translation and rotation invariant and generated only roads that initially point to north.
This was not the case in the competition.

Frenetic 2021 \cite{frenetic} is a genetic algorithm with domain-specific mutations which are applied only to failing
tests in order to improve solution diversity. Once the algorithm finds a failing test it can mutate it, e.g., by
mirroring. This produces new tests that are likely to fail and are diverse by construction. It is possible to extend
WOGAN to include such mutations using domain-specific rules. We conjecture this would increase WOGAN's failure-inducing
test diversity metric. We also conjecture that a more varied initial random search could lead to more diverse failing
tests.

\begin{acks}
We thank the organizers of the SBST competition for creating and continuing the competition.

This research work has received funding from the ECSEL Joint Undertaking (JU) under grant agreement No 101007350.
\end{acks}

\bibliographystyle{ACM-Reference-Format}
\bibliography{sbst}

\end{document}